\documentclass[11pt]{article}

\usepackage{iclr2023_conference}
\iclrfinalcopy
\usepackage{times}

\usepackage{times}
\usepackage{array, makecell}
\usepackage{latexsym}
\usepackage{hyperref}

\usepackage[T1]{fontenc}

\usepackage[utf8]{inputenc}

\usepackage{microtype}
\usepackage{booktabs}
\usepackage{listings}
\usepackage{caption}
\usepackage{multirow}
\usepackage{times}
\usepackage{latexsym}
\usepackage{enumitem}
\usepackage{amsmath,amssymb,amsfonts}
\usepackage{xcolor}

\newcommand{\monourl}[1]{\href{#1}{\texttt{#1}}}

\definecolor{lightred}{rgb}{1, 0.7, 0.7}
\definecolor{lightgreen}{rgb}{0.7, 1, 0.7}

\definecolor{codegreen}{rgb}{0,0.6,0}
\definecolor{codegray}{rgb}{0.5,0.5,0.5}
\definecolor{codepurple}{rgb}{0.58,0,0.82}
\definecolor{backcolour}{rgb}{0.95,0.95,0.92}
\definecolor{lightyellow}{rgb}{0.98, 0.91, 0.71}

\lstdefinestyle{mystyle}{
    backgroundcolor=\color{white},   
    commentstyle=\color{codegreen},
    keywordstyle=\color{magenta},
    numberstyle=\tiny\color{codegray},
    stringstyle=\color{codepurple},
    basicstyle=\ttfamily\footnotesize,
    breakatwhitespace=false,         
    breaklines=true,                 
    captionpos=b,                    
    keepspaces=true,                 
    numbersep=5pt,                  
    showspaces=false,                
    showstringspaces=false,
    showtabs=false,                  
    tabsize=2,
    escapechar=|,
}
\makeatletter

  \newcount\bt@rangea
  \newcount\bt@rangeb

  \newcommand\btIfInRange[2]{%
      \global\let\bt@inrange\@secondoftwo%
      \edef\bt@rangelist{#2}%
      \foreach \range in \bt@rangelist {%
          \afterassignment\bt@getrangeb%
          \bt@rangea=0\range\relax%
          \pgfmathtruncatemacro\result{ ( #1 >= \bt@rangea) && (#1 <= \bt@rangeb) }%
          \ifnum\result=1\relax%
              \breakforeach%
              \global\let\bt@inrange\@firstoftwo%
          \fi%
      }%
      \bt@inrange%
  }
  \newcommand\bt@getrangeb{%
      \@ifnextchar\relax%
          {\bt@rangeb=\bt@rangea}%
          {\@getrangeb}%
  }
  \def\@getrangeb-#1\relax{%
      \ifx\relax#1\relax%
          \bt@rangeb=100000
      \else%
          \bt@rangeb=#1\relax%
      \fi%
  }

  \newcommand{\btLstHL}[2]{%
    \btIfInRange{\value{lstnumber}}{#1}{#2}{}
  }%

   \makeatletter
   \let\old@lstKV@SwitchCases\lstKV@SwitchCases
   \def\lstKV@SwitchCases#1#2#3{}
   \makeatother
   \usepackage{lstlinebgrd}
   \makeatletter
   \let\lstKV@SwitchCases\old@lstKV@SwitchCases
   
   \lst@Key{numbers}{none}{%
       \def\lst@PlaceNumber{\lst@linebgrd\ }%
       \lstKV@SwitchCases{#1}%
       {none:\\%
        left:\def\lst@PlaceNumber{\llap{\normalfont
                   \lst@numberstyle{\thelstnumber}\kern\lst@numbersep}\lst@linebgrd}\\%
        right:\def\lst@PlaceNumber{\rlap{\normalfont
                   \kern\linewidth \kern\lst@numbersep
                   \lst@numberstyle{\thelstnumber}}\lst@linebgrd}%
       }{\PackageError{Listings}{Numbers #1 unknown}\@ehc}}
   \makeatother

\definecolor{l1}{HTML}{1b9e77}
\definecolor{l2}{HTML}{d95f02}
\definecolor{l3}{HTML}{7570b3}
\definecolor{ll1}{HTML}{e7fef7}
\definecolor{ll2}{HTML}{feede6}

\newcommand{\inlinecode}{\texttt}

\newcommand{\github}{https://anonymous.4open.science/r/BigIssue-8EE2/}

\title{BigIssue: A Realistic Bug Localization Benchmark}

\author{%
  Paul Kassianik\thanks{Equal Contribution} \\
  Salesforce Research\\
  \texttt{pkassianik@salesforce.com} \\
  \And 
  Erik Nijkamp\footnotemark[1] \\
  Salesforce Research \\
  \texttt{erik.nijkamp@salesforce.com} \\
  \And
  Bo Pang \\
  Salesforce Research \\
  \texttt{bo.pang@salesforce.com} \\
  \And
  Yingbo Zhou \\
  Salesforce Research \\
  \texttt{yingbo.zhou@salesforce.com } \\
  \And 
  Caiming Xiong \\
  Salesforce Research \\
  \texttt{cxiong@salesforce.com} \\
}

\begin{document}
\maketitle
\begin{abstract}
As machine learning tools progress, the inevitable question arises: How can machine learning help us write better code? With significant progress being achieved in natural language processing with models like GPT-3 and BERT, the applications of natural language processing techniques to code is starting to be explored. Most research has been focused on automatic program repair (APR), and while the results on synthetic or highly filtered datasets are promising, such models are hard to apply in real-world scenarios because of underperforming bug localization techniques. We propose BigIssue: a realistic bug localization benchmark. The goal of the benchmark is two-fold. We provide (1) two general benchmarks with a diversity of real and synthetic Java bugs and (2) a motivation to improve bug localization capabilities of models through longer context encodings. With the introduction of BigIssue, we hope to advance the state of the art in bug localization, in turn improving APR performance and increase its applicability to the modern development cycle.
\end{abstract}

\section{Introduction}

Recent advances in natural language processing (NLP) \citep{brown2020gpt3, devlin2018bert, liu2019roberta} have increased interest in applying NLP techniques to code understanding. With the development of code encoders \citep{codebert, cubert}, this task is becoming increasingly more accessible and appealing. As research has jumped ahead into the task of Automated Program Repair (APR), the results have been not been adequate. Although synthetic datasets have largely been solved (see Section \ref{synthetic_benchmarks}), models have been surprisingly underperforming on real-world datasets, many not even able to repair a quarter of the bugs in the Defects4J benchmark (see Section \ref{sbfl} and \citet{coconut}). This is despite research suggesting that current APR benchmarks suffer from a lack of diversity  \citep{durieux2019empirical}. As a consequence, many APR models are prone to overfitting to specific datasets \citep{apr_survey_obstacles}. Although interesting from an academic perspective, such tools would hardly be useful in a real industrial scenario.

We posit that the three major limitations to APR methods being used today are: (1) training to fix already located bugs rather than finding bugs and fixing them, (2) the inability of models to take large contexts into account, and (3) the reliance on information besides pure code. The first limitation is straightforward: patches have limited context outside of the lines immediately before and after each patch. It has been shown that APR performance improves significantly if a good fault localization algorithm is used to detect buggy code locations \citep{durieux2019empirical, liu2019you}. The second limitation prevents models from finding bugs that depend on the context of the program. Even for human readers many real-world bugs require a lot of program-specific context to be detectable. One of the most popular code encoders today \citep{codebert} only supports encoding of sequences up to 512 tokens, not nearly enough to process most Java files in real-world programs (on average 7.5k tokens with the RoBERTa tokenizer \citep{liu2019roberta}). The third limitation follows from the fact that one of the most commonly bug localization methods, SBFL \citep{jones2005empirical}, is heavily reliant on test cases exposing potentially buggy locations.

In order to advance the state of the art of both BL (Bug Localization) and APR (Automatic Program Repair) models, we introduce BigIssue. The major contributions of BigIssue include: 
\begin{itemize}
\item A large collection of confirmed real-world bugs with line-level annotations. Each bug has been reported by live users to the GitHub Issues bug-tracking system and fixed via a commit or pull request. The dataset contains a total of $10,905$ bugs sourced from $4,233$ Java repositories. 
\item A long-sequence synthetic bug dataset. Perturbations in real code collected from GitHub are generated by InCoder \citep{fried2022incoder}, a state-of-the-art code generation model.
\item An empirical demonstration of the hardness of the real benchmark as compared to a synthetic benchmark. Even with advanced synthetic bug generation techniques, the performance on real bugs of models trained on synthetic data will not be adequate, which calls for further research into realistic bug detection.
\end{itemize}

By providing a large and diverse dataset of synthetic and real bugs from a multitude of projects without any extra information outside of code, we hope to push the direction of research towards line-level long-context bug localization for better performance on APR tasks.

\section{Prior Art} \label{prior_art}

\subsection{Automatic Program Repair}

Since bug localization is fundamentally related to automatic program repair, we provide a brief survey of existing APR benchmarks and their drawbacks.

\paragraph{Real-world Benchmarks}

The Defects4J dataset by \citet{defects4j} has been widely used in automatic program repair. It consists of 357 (835 is version 2) bugs sourced from 10 (100) top open-source Java projects. Bugs are manually reviewed and each bug has at least 1 test case that exposes the bug. APR methods, however, are not successful enough on Defects4J to suggest utility in real-world applications. The most recent state of the art model can only fix 67 out of 357 bugs \citep{circle}, while the two previous state of the art models could only fix 44 \citep{coconut} and 57 \citep{jiang2021cure} bugs. This is despite recent research that suggests APR methods are overperforming on Defects4J as compared to other similar benchmarks \citep{durieux2019empirical}.  Bugs.jar \citep{saha2018bugs} is a similar dataset but with an expanded scope of 8 popular projects from the Apache foundation.

iBugs \citep{dallmeier2007ibugs} presents a methodology of semi-automatic bug extraction from repositories, and provides a concrete dataset of applying the methodology to the ASPECTJ project. The method involves analyzing commit logs for signs that indicate a bug fix, extracting the pre-commit and post-commit versions of the repositories, and identifying test cases that represent the bug-fix. However, this dataset is fairly small and is only sourced from one repository.

Another widely used dataset is the \texttt{ManySStubs4J} dataset \citep{karampatsis2020often}. It's a collection of many ``stupid" bugs mined from 100 (1,000) top open-source Java repositories. The collection includes only those changes where the change is a single line of code and falls into one of pre-determined 16 categories of bugs. While convenient, it suffers from a lack of complicated bugs and highly selective criteria.

Learning-fixes \citep{tufano2018empirical} is a collection of about 58,350 short methods mined from GitHub. Each of the methods was semantically idiomized and presented in the benchmark. The main limitation of this dataset is that it's a method-level dataset: each bug should be identifiable and fixable based on the context only present in that particular method. For real bugs, this is usually not the case.

DLFix \citep{li2020dlfix} is another dataset aimed at APR tasks. The dataset consists of almost 5 million methods, enhanced with metadata, and the corresponding fixed versions of the method for a particular repository. While interesting for limited cases, the method-level granularity as well as the necessity of building metadata for each method limits its usefulness, especially on longer methods. 

Table \ref{table:comparison} presents a comparison of existing APR benchmarks.

\paragraph{Synthetic Benchmarks} \label{synthetic_benchmarks}

A natural way to deal with the lack of data diversity in current real-world benchmarks is to create synthetic benchmarks by perturbing code. The simplest way to create code perturbations is to apply rule-based perturbations to a corpus of code \citep{cubert} or via a static oracle (such as a linter) \citep{berabi2021tfix}. Other datasets are generated via a separate perturbation model. SPoC \citep{kulal2019spoc} uses a simple LSTM to generate lines of code that might be potentially buggy. DeepDebug \citep{drain2021deepdebug} uses a more complicated model trained on reversed git commits to generate synthetic bugs. While attractive, there is significant evidence that good performance on these benchmarks does not translate to good performance on real-life bugs \citep{durieux2019empirical}. We also perform experiments in Section \ref{findings} that suggests that even good performance on sophisticated perturbation datasets does not translate well to fixing real bugs.

\subsection{Using Existing Benchmarks for Bug Localization} \label{sbfl}

Line-level fault localization and fault prediction on their own have been severely understudied. While file-level localizations can achieve up to 70\% top-5 accuracy scores \citep{lam2017bug}, line-level localization methods aren't as successful. According to a recent survey by \citet{zou2019empirical} current fault localization and prediction methods can't even localize half of the bugs in the Defects4J \citep{defects4j} dataset. The most widely used method for fault localization is Spectrum-based fault localization (SBFL) \citep{jones2005empirical}. While elementary and simple to implement, it relies heavily on the quality and quantity of test cases, especially for large programs \citep{keller2017critical}. Newer deep learning methods such as TRANSFER-FL \citep{meng2022improving} also can't consistently localize lines of buggy code (only 84/395 examples). 

\begin{table*}[htbp]
    \begin{center}
    \small
    \begin{tabular}{p{11em} c l l l l l}
         \toprule
         Dataset & Size & Granularity & Bug Length & Context & \# of Repos & Filters \\
         \midrule
         BigIssue & 10,905 & Line & Multi-line & Repo & 4233 & No\\
         \midrule
         Defects4J\citep{defects4j} & 357 (835) & Line & Multi-line & Repo & 5 (17) & No \\
         Bugs.jar\citep{saha2018bugs} & 1158 & Line & Multi-line & Repo & 8 & No \\
         \multirow{3}{11em}{ManySStubs4J \citep{karampatsis2020often}} & 10,231  & \multirow{3}*{Line} & \multirow{3}*{Single-line} & \multirow{3}*{Repo} & 100 & \multirow{3}*{Yes} \\ 
         & (63,923) & & & & (1000) & \\
         & & & & & & \\
         iBugs \citep{dallmeier2007ibugs} & 369 & Line & Multi-line & Repo & 1 & No \\
         Learning-Fixes \citep{tufano2018empirical} & 58,350 & Line & Multi-line & Method & - & No \\
         DLFix \citep{li2020dlfix} & 4,973,000 & Method & Multi-line & Repo & 8 & No \\
         \bottomrule
    \end{tabular}
    \vspace{8pt}
    \caption{Comparison of Major Java Bug Detection Datasets.}
    \label{table:comparison}
    \end{center}
    \medskip
\end{table*}

\section{BigIssue Synthetic Dataset}





\subsection{Motivation}

Evaluation of approaches towards bug localization requires the construction of a dataset with known ground-truth bugs. One methodology to create such dataset is to consider existing code and introduce erroneous perturbations in the form of samples drawn from a generative model. In prior art \citet{kulal2019spoc}, synthetic perturbations have been adopted on a function-level granularity with weak generative models such as small LSTMs. The underlying distribution of such synthetic dataset may be quite dissimilar to the distribution of realistic bugs, which occur in software engineering \citep{durieux2019empirical}. To decrease this discrepancy, we will advance this concept to file-level data and sample perturbations from a strong generative model.


Our synthetic dataset adopts the methodology of gathering ``real'' code as observations and introducing synthetic perturbations in the observations. Here, the perturbation is a rewrite of the original sequence of code into a perturbed sequence of code. In our approach, a portion of the original code is ``masked out'' and a generative model is recruited to ``fill in'' the masked out code. The ``filled in'' portion of the code constitutes the synthetic perturbation. The perturbation of the original code is assumed to likely to contain ``errors''.


While the above approach based on perturbations may appear obvious and trivial, the construction of such datasets is challenging. This is due to the following conditions: (1) existing code is not guaranteed to be free of errors, (2) the definition or ontology of an ``error'' or ``bug'' itself is non-trivial, (3) the creation of synthetic perturbations that are difficult to discriminate from original observations and yet reflect the distribution of ``real'' errors is hard.

Prior art addresses these issues by (1) reducing the scope of the code to function or line-level, effectively reducing the span of code to $n$ lines of code \citep{cubert, yasunaga2020drrepair, yasunaga2021break}, (2) introducing heuristic perturbations rules or pre-defining a set of categories in which ``bugs'' fall \citep{cubert, drain2021deepdebug}, or (3) perturbing a single line of code in simple programs \citep{yasunaga2020drrepair, drain2021deepdebug}. While this over-simplification is a reasonable first step, the resulting dataset may be quite different from realistic errors for which localization is deemed ``useful'' to a practitioner.

Our work addresses (1) and (2) by doing away with the notion of an ``error'' and instead shifting the conceptual thinking towards the distributions of ``original'' and ``perturbed'' observations. That is, our dataset is assumed to contain errors which are not identified in the ground-truth labels. The task of error localization is relaxed as the task of localization of perturbations. This relaxation allows us to consider file-level observations without the need for a strict definition of an ``error''. Such relaxed definitions is suitable the construction of a ``sanity-check" dataset to test a model's capability of detecting code divergent from standard coding practices. In the following, we will provide details on the creation of such dataset and in particular address (3).

\subsection{Dataset Construction}

The underlying methodology of the creation of this dataset is (1) to gather large amounts of file-level observations (i.e., real code), (2) to introduce synthetic perturbations from a strong generative model such that discrimination of ``original'' and ``perturbed'' observation is non-trivial, (3) and relax the task of ``error localization'' to the task of ``perturbation localization''. In this section, we describe the construction of such a dataset.


\paragraph{Observations}
In order to obtain large quantities of observations for the learning and evaluation of localization models, the proposed dataset is a compilation of public, non-personal information from GitHub consisting of permissively licensed Java code in October 2021. In particular, we gathered 8 million repositories between January 2014 and October 2021 annotated with at least 1 star and considered the subset of contained files containing Java code. The files must have an average length of $\le100$ characters and a maximum line length of $1,000$. Files where $\ge90\%$ of the characters are decimal or hexadecimal digits are also removed. Finally, exact duplicates based on their SHA-256 hash are removed, which amounts to a substantial portion of the raw data due to forks and copies of repositories. The resulting dataset comprises 96.56 GB of raw text.


\paragraph{Perturbations}
For realistic perturbations, we resort to a method known as ``inpainting'' for images or ``infilling'' for the textual domain. That is, a portion of a giving observation is occluded (or masked out). Then, the occlusion is reconstructed or ``filled in'' by a sample drawn from a generative model conditional on the non-occluded context. Recently, auto-regressive causal language models, such as \citet{brown2020gpt3}, have demonstrated to excel at this task for which the prompt may be treated as context and the auto-regressive sample conditional on the prompt as the in-painting while preserving the statistical regularities of the training data. However, the joint distribution over tokens is usually factorized in a left-to-right order over time, for which the causal mask constraints the infill samples to only take past context into account, but not future tokens. In our case of sampling realistic perturbations at random spans within a given observation, we wish to take both the code before and after the masked out span into account for file-level consistency. To address this issue, we recruit an auto-regressive sampler that re-arranges the input sequence and associated causal masking such that sampling is conditional on both past and future context \citep{du2022glm, fried2022incoder}. To further reduce the gap between ``real'' and ``perturbed'' sequences, we chose a large-scale language model, InCoder \citep{fried2022incoder}, with 1~billion parameters, and lowered the temperature of auto-regressive nucleus sampling to $0.8$. This temperature value was selected by manual experimentation. Equipped with such a sampler, a random span in the observation is removed and infilled with a sample drawn from the InCoder model. The length of the span is drawn from a uniform distribution with minimum length of $64$ tokens and maximum length of $128$ tokens. The generated sample is constrained to at most the length of the span.


To further improve the quality of perturbations, we use rejection sampling from the InCoder model where drawn samples not satisfying the formal grammar of the programming language are rejected. Specifically, we (1) reject any files which are not syntactically correct \footnote{To verify syntactical correctness of Java programs, we use the \textsc{JavaLang} library: \monourl{https://github.com/c2nes/javalang}. }, (2) reject files containing less than $2,048$ tokens, (3) reject perturbations for which $10$ attempts of infill sampling (with a minimum span length of $64$ and a maximum number of tokens of $128$) did not result in a syntactically correct perturbation, (4) reject samples for which the Levenshtein distance between the unperturbed and perturbed sequence is smaller than $64$ tokens or larger than $192$ tokens.

\paragraph{Task}
Our proposed ``perturbation localization'' task can be expressed in the form of a binary classification for which each line is labeled as either ``original'' or ``perturbed''. As such, the ground-truth labels indicate whether the line is a sub-sequence of the observation or was (potentially partially) perturbed by the sampler. Each file contains at most one such perturbation. The length of the input sequence is limited to at most $8,192$ tokens under the RoBERTa tokenizer \citep{liu2019roberta} with at most $512$ lines per file.

\subsection{Dataset Examples and Artifacts}

Some samples from the synthetic dataset are presented in Appendix \ref{appendix:synthetic_samples}, and all artifact details can be found in Appendix \ref{appendix:data_desc}.



\section{BigIssue Realistic Benchmark}

\subsection{Motivation}

Based on our observations about existing benchmarks from Section \ref{prior_art}, we concluded that a new benchmark is needed to push the state of the art forward. Therefore, we created a benchmark that prioritized quantity over perceived quality and one that focused specifically on NL-based line-level bug localization. 

For this benchmark, we defined a line as ``buggy" if it has been removed or modified in the issue patch. This allows us to avoid using tests as the ground truth for bugs in code. This definition also fits well with the usage of code encoders such as CodeBERT \citep{codebert} for line-level classification, as demonstrated in Section \ref{findings}.

\subsection{Benchmark Construction}

First, we considered Java GitHub repositories created between January 2014 and October 2021. In order to ensure that we only filter out repositories that were intended for some form of public use, we only examined repositories with at least 1 star. We further filtered down the repositories to only those repositories that had GitHub Issues enabled and had licenses permitting use of their code (full list of licenses is available in Appendix \ref{appendix:ethical}). That gave us $4,233$ repositories.

Using the GitHub API we filtered through closed issues on these repositories. We only used public, non-personal information available through the API. In order to select issues that corresponded to bug fixes on that particular repository, we selected issues that either contained ``bug", ``fix", or ``fixed" as separate words in the title and the body of the issue. We also included issues that contained the label ``bug".  We looked at issues with a corresponding ``close" event, and we looked at the commit that was attached to the latest "close" event. This gave us a dataset of $23,924$ total closed issues. We further filter only those bugs that affect one Java file without test code. That yields $10,905$ bugs.

To verify the validity of our filters, we manually verified 100 sample issues. We manually verified the validity of 84 of the issues. A detailed breakdown can be found in Appendix \ref{appendix:realistic_analysis}.


Similarly to iBugs \citep{dallmeier2007ibugs}, to identify buggy lines we examine the data from the hunks in the diff. If a line is (1) removed from the source file and (2) is not an import line (lines that begin with \texttt{import ...}), it is marked as buggy. In cases where hunks are exclusively adding code, we mark the two lines in the source before and after the change as buggy. Processing added code is not always straightforward: sometimes the added chunk is an outsourced piece of code from a different method. However, this simplification of the process was done to account for added code while minimizing the potential impact of simply outsourced chunks.

\paragraph{Test-running frameworks} Many of the benchmarks presented above use tests either as assistance in bug fixing or as a method of filtering bugs. We do not consider testing frameworks and tests as criteria for whether a commit is a bug or not. Firstly, it was recently shown that unit tests on their own do not guarantee fewer failures inside the code \citep{chioteli2021does} which implies that there are even more bugs inside the code that are not exposed by tests. Secondly, we would be severely limiting the diversity and scope of our benchmark by forcing issues to include an exposing test case.

\subsection{Benchmark Examples and Artifacts}

Some samples from the synthetic dataset are presented in Appendix \ref{appendix:realistic_samples}, and all artifact details can be found in Appendix \ref{appendix:data_desc}.

\section{Synthetic vs Realistic Bug Detection} \label{findings}


In this Section, we conduct a preliminary analysis of the hardness of the BigIssues benchmark. Since the sequence length exceeds the limitations of most pre-trained language models on code, we recruit mean pooling to construct simple baselines. 
We hypothesize that although the realistic data is much harder than the synthetic dataset, using long-context encoders in addition to synthetic pre-training will help increase performance.

\subsection{Hypothesis}

The proposed BigIssue benchmark contains two variants: (1) synthetic rewrites of real code sampled from a strong generative model, (2) realistic rewrites of real code based on the commits associated with a closed issue in GitHub.

Recall, for (1) a recent large language model was recruited as a sampler which, compared to prior art, not only is of significant size under scaling laws, but furthermore alters the causal masking such that future tokens can be taken into account as context. We argue that these synthetic rewrites are non-trivial to detect compared to prior art.

However, our hypothesis is that localization of real bugs is still a significantly harder task not solvable with elementary models and requires substantial research to be solved. While local, trivial bugs do not require context to be localized, harder non-local bugs can often only be resolved when taking the entire file, a set of imported files, or the entire repository into account. We use elementary models to show that (a) synthetic bugs are easier to detect than real bugs, and can serve as a sanity check for bug localization models and (b) we use synthetic bugs to show that longer-context encodings improve performance on bug localization.


\subsection{Model}


Our architecture partitions a long input sequence of $8,192$ tokens into shorter sub-sequences, computes contextualized vectors for each chunk using a bi-directional encoder model, combines the contextualized vectors into $512$ latent vectors with mean-pooling, and finally projects those vectors to logits for line-level binary classification. 

Consider a sequence $x = (x_0, x_1, \ldots, x_n)$ of input tokens with length $n=8,192$. To address the issue (2) of large $n$, we partition $x$ into $m=16$ equally sized chunks $\tilde{x}_i$ with $i\in\{0,\dots,15\}$ each containing $512$ tokens. To contextualize the embedding vector of the tokens, we recruit the pre-trained bi-directional encoder $f$, (such as CodeBERT \citep{feng2020codebert}), and compute $f(\tilde{x}_i)$ for each partition $i$. Then, the contextualized partitions are concatenated $\hat{x} = (f(\tilde{x}_0), f(\tilde{x}_1), \ldots, f(\tilde{x}_m))$. To restore global position information, we apply additive sinusoidal positional embeddings to $\hat{x}$. A layer of self-attention integrates the information across partition boundaries. Mean-pooling is applied to $\hat{x}$ with a window length such that the resulting sequence of latent vectors matches the maximum number of $512$ lines. A standard linear projection maps each of the line-level latent vectors to logits for binary classification. The resulting model is fine-tuned with binary cross entropy as the objective function.

The appeal of the proposed model is to leverage the representations learned by a strong backbone model and the simplicity in handling variable length including line breaks in the input sequence. CodeBERT has demonstrated strong empirical performance on downstream tasks so the learned representations should be well suited for bug localization. To demonstrate the utility of long context for code understanding, we also use the standard Longformer \citep{beltagy2020longformer} as an encoder. The mapping of contextualized vectors to latent vectors allows for variable length input sequences and avoids special treatment of newline characters. The alignment from lines of the input sequence to latent vectors for classification is implicitly learned by supervision.

\begin{table*}[!h]
\small
\begin{center}

\begin{tabular}{p{2cm}ccp{0pt}ccp{0pt}cc}
\toprule
\multirow{2}{*}{Model} & \multicolumn{2}{c}{Recall$^\uparrow$} & & \multicolumn{2}{c}{Precision$^\uparrow$} & & \multicolumn{2}{c}{F1$^\uparrow$}\\\cmidrule{2-3}\cmidrule{5-6}\cmidrule{8-9}
& Synthetic & Realistic & & Synthetic & Realistic & & Synthetic & Realistic \\\midrule
Random & 49.58 & 50.99 & & 2.68 & 0.96 & & 5.08 & 1.88\\
Pooling & 93.48 & 69.43 & & 8.89 & 2.16 & & 16.24 & 4.17\\
Pooling-Attn & 95.37 & 64.66 & & 26.93 & 1.84 & & \textbf{42.00} & 3.58\\
\bottomrule
\end{tabular}
\end{center}
\caption{Comparison of the binary classification accuracy under various baselines: (1) Random Bernoulli classifier with $p=0.5$, (2) Mean pooling model, (3) Mean pooling model with self-attention between latent vectors.}
\medskip
\label{tab:f1}
\end{table*}

\begin{table*}[!h]
\small
\begin{center}
\begin{tabular}{p{3cm}cccp{0pt}ccp{0pt}cc }
\toprule
\multirow{2}{*}{Model} & \multirow{2}{*}{Training} & \multicolumn{2}{c}{Recall$^\uparrow$} & & \multicolumn{2}{c}{Precision$^\uparrow$} & & \multicolumn{2}{c}{F1$^\uparrow$}\\\cmidrule{3-4}\cmidrule{6-7}\cmidrule{9-10}
& & Synth. & Real. & & Synth. & Real. & & Synth. & Real. \\ \midrule
Longformer-4096 & Synthetic & 98.49 & 42.98 & & 22.62 & 3.74 & & \textbf{36.79} & 6.88 \\
Longformer-512 & Synthetic & 97.54 & 46.44 & & 18.78 & 3.94 & & 31.50 & 7.27 \\
\midrule
Longformer-4096 & Real  & 73.28 & 75.40 & & 5.92 & 2.65 & & 10.96 & 5.12 \\
Longformer-512 & Real  & 81.28 & 88.68 & & 5.95 & 2.46 & & 11.09 & 4.79 \\

\end{tabular}
\end{center}
\caption{Comparison of Longformer Models. The numbers show that synthetic training is a suitable proxy task for realistic bug detection compared to exclusively realistic training and the advantages of long-context on synthetic data. 
}
\medskip
\label{tab:longformer}
\end{table*}

\subsection{Findings}

To evaluate the hardness of the artifical and realistic BigIssue benchmark, the aforementioned model is trained on both datasets.  
Training details can be found in Appendix \ref{appendix:training_details}.

Table~\ref{tab:f1} summarizes the binary classification performance in terms of recall, precision and F1-score for three baseline models with CodeBERT encoder: (1) A random classifier for which the line-level predictions are modeled as a Bernoulli random variable per line with probability $p=0.5$, (2) a mean-pooling based model for which the self-attention layer between latent vectors is omitted, (3) a mean-pooling based model including self-attention between latent vectors.

For the synthetic dataset, the mean-pooling model including self-attention with an F1-score of $42.00$ significantly improves over the random Bernoulli baseline with $5.08$. Self-attention to integrate information across latent vectors improves the score by nearly $26$ points, which may indicate that attention across the partitioning of $512$ tokens is crucial. One may assume with further improvements in modeling, the synthetic dataset is solvable.


For the realistic benchmark, however, both of the mean-pooling baselines performed better than random. The extra layer of attention did not add any improvement. Both models tend to have high recall values, but precision is especially low for the realistic benchmark.

To test the effect of longer-context encoders, we replaced the encoder in our Mean-Pooling with Attention model with a Longformer \citep{beltagy2020longformer} that is capable of handling sequences up to $4096$ tokens. Instead of chunking the sequence into 16 chunks of $512$, we chunked it into 2 chunks of $4096$. We trained a Longformer-4096 token Mean-Pooling with Attention model. We also trained the same model solely on realistic data. The results after $50,000$ steps of training are presented in Table~\ref{tab:longformer}. The results on synthetic data suggest that longer-context encoders improve performance. While the 512-chunked model performed better than the 4096-chunked model on realistic data, the difference cannot be described as significant due to low precision values.

As hypothesized, real bug detection is a much harder challenge than synthetic data. Simple models could not solve the realistic bugs with the same success as the synthetic bugs, showing that synthetic bugs are a suitable sanity check for bug localization models. The findings also suggest that using longer contexts is effective for catching synthetic bugs. It is our hope that this finding spurs research toward the modeling of long contexts to approach the task of real bug detection.

\section{Conclusion}

We propose a new realistic benchmark for line-level bug localization that does not rely on test suites. We also provide a synthetic benchmark and dataset, generated with more sophisticated methods than previous work. We also show that despite the discrepancy between synthetic and realistic bugs, synthetic benchmarks can be used as a sanity check for bug localization models. Using the synthetic dataset, we show that long-context encodings help in bug localization, and we hope that these results push research into long-context for realistic bug localization.

We hope that our contributions inspire and push future research into realistic, long-context, NLP-based bug localization techniques. Advances in this area would bring automatic program repair to a state that would be useful and transformative to the modern software development process.



\bibliography{refs}
\bibliographystyle{iclr2021_conference}



\UseRawInputEncoding

\appendix

\section{Data Description, Hosting Details, and Data Access}
\label{appendix:data_desc}

We publish the training, evaluation, and validation sets for the synthetic data. We also publish the realistic benchmark. These items can be accessed in a Google Cloud Storage bucket at \monourl{https://console.cloud.google.com/storage/browser/bigissue-research}. All materials are released under the MIT License. 

\paragraph{Realistic Pre-training data} For the realistic Pooling and Pooling-Attention models, we created a pre-training dataset similar to other projects. We select Java GitHub repositories with 5 stars or more, we clone the main branch of the repository, while only downloading files under 2 megabytes. We then filter the commits that include the words "error", "bug", "fix", "issue", "mistake", "incorrect", "fault", "defect", "flaw", or "type", using standard practice in ManySStubs4J project \cite{karampatsis2020often} . Since our models are designed only for single-file bug localization, we take each modified file and apply the labeling procedure described in the paper to generate the examples and labels. We truncate files at 8192 tokens in the same manner as in \cite{codebert}. In total, we get about 195 GB of data to use for pre-training.

\section{Training Details}
\label{appendix:training_details}

We train all of our models on a single pod with 16 A100 GPUs. For models with the CodeBERT encoder, we optimized the model with a linear schedule AdamW \cite{loshchilov2017decoupled} optimizer, with a starting learning rate of 5e-5, and $10,000$ warmup steps. We train over 50,000 steps with a batch size of 8. For models with the Longformer encoder, we optimized with a linear schedule AdamW optimizer, starting learning rate of 3e-5, and $1000$ warmup steps.

We provide the full training code at \url{\github}. 

\paragraph{Model Checkpoints} We provide the model checkpoints for the Pooling and Pooling-Attention models trained on realistic data in the GitHub repository \url{\github}.

\section{Data Collection Ethical Statement}
\label{appendix:ethical}

We did not collect any personal information from the GitHub API. We only collect commit information and data inside the commits, without taking into the account the origin or the user profile of the user making the changes. 

We also present here the list of licences that we use in our paper: \url{\github licences.txt}

\section{Realistic Bug Analysis}
\label{appendix:realistic_analysis}

We've analyzed the quality of our selection process by sampling 100 random issues. This will allow other researchers to use this method with greater flexibility in licensing. We analyzed them based on three criteria: (a) whether the issue represents a valid bug (b) whether the fix represents a valid fix for the bug and (c) whether the fix depends on code conventions or APIs, and is therefore identifiable by a human with adequate knowledge of the above. The last criteria is to ensure possible identifiability by a model: if a bug is only marked as such because of outside software criteria, there is no reasonable way a model can learn that bug pattern.

In total, we have collected: 
\begin{itemize}
    \item 84 valid issues.
    \item 3 issues that did not represent valid bugs.
    \item 3 issues were the fixes did not fix the bugs.
    \item 2 issues that were not subject to analysis.
    \item 8 issues that are not identifiable without outside context.
\end{itemize}

Each item in the following list represents an issue considered for our dataset and is identified by the corresponding link to the github issue. We precede invalid issues with "*" for clarity.

\begin{enumerate}
    \item * - \url{https://github.com/gsantner/markor/issues/314} - Valid bug, but the fix doesn’t address the issue. In fact the issue persists despite the issue being closed.
 
    \item \url{https://github.com/ReplayMod/ReplayMod/issues/422} - Valid bug, valid fix for the bug, bug is identifiable through similar usage of code around the buggy area.
 
    \item \url{https://github.com/lingochamp/FileDownloader/issues/855} - Valid bug, valid one-line fix for the file, bug is identifiable because `model.status` is being set in almost all action methods of this class.
 
    \item \url{https://github.com/danielCantwell/Fit-Friend/issues/3} - Valid bug, valid fix. Bug is identifiable by the fact that the variables used in that particular area of code are not used at all.
 	
    \item \url{https://github.com/OpenJEVis/JEVis/issues/1699} - Valid bug, valid fix. Bug is an edge case if the JEVis samples are not sampled at a regular interval. Bug is identifiable because of similarities in time-series interactions.
 
    \item \url{https://github.com/TechReborn/TechReborn/issues/549} - Valid bug, valid fix. Bug is identifiable because (a) a lot of values are hard-coded and (b) some of the variables did not follow the human logic of what a minecraft “operator” is supposed to do (e.g. not checking if there is any space in the output container for more items to go to).
 
    \item \url{https://github.com/milaboratory/mixcr/issues/509} - Valid bug, valid fix, although there are extraneous style fixes. The style fix is that whenever the “percentage” is being used, the text for the log must be written as “percent used …”. This follows a pattern from other places in this repository. The bug is that the percentages can sometimes be over 100
 
    \item * - \url{https://github.com/Angry-Pixel/The-Betweenlands/issues/895} - Valid bug, and valid fix, but the bug is not generally identifiable by humans without external context. The minecraft bug suggests that a certain item has to be not repairable, but from the code alone there seems to be no suggestion that this item should be unrepairable.
 
    \item \url{https://github.com/VazkiiMods/Quark/issues/2920} - Valid bug, valid fix. Identifiable by humans if they have knowledge and context about the Create library that cannot accept \texttt{FakePlayers} as players when performing operations on the world.
 
    \item \url{https://github.com/15knots/cmake4eclipse/issues/26} - Valid bug, valid fix. The essence of the bug is that certain build directories might be deleted outside of Eclipse, and that needs to be handled by the code, which is a common safeguard that code needs to implement.
 
    \item \url{https://github.com/TheThirdOne/JSoftFloat/issues/1} - Valid bug, valid fix. This bug should be identifiable with knowledge of floating-point arithmetic calculations.
 
    \item \url{https://github.com/ververica/flink-cdc-connectors/issues/326} - Valid bug, valid fix. To identify this bug, one must need to know that a certain parameter in the config can be null. In particular, if there is no pre-defined database history ID, one must be set.  
 
    \item \url{https://github.com/spring-cloud/spring-cloud-sleuth/issues/333} - Valid bug, valid fix. Bug is identfiable based on existing usage from other Callable services, where the class is frequently used as a wrapper for calls.
 
    \item \url{https://github.com/tango-controls/rest-server/issues/192} - Valid bug, valid fix. Identifiable with knowledge of the TANGO API specification.
  
    \item \url{https://github.com/guillaume-alvarez/ShapeOfThingsThatWere/issues/7} - Valid bug, valid fix. The issue is that the camera movement logic is tied to frame-processing, and if that pattern is known then the bug is identifiable.
  
    \item \url{https://github.com/spring-cloud/spring-cloud-config/issues/128} - Valid bug, valid fix. Bug identifiable by common environment loading patterns.
  
    \item \url{https://github.com/BentoBoxWorld/TwerkingForTrees/issues/9} - Valid bug, valid fix. Identifiable by minecraft logic of not allowing players to modify blocks outside of world border.
 
    \item \url{https://github.com/requery/requery/issues/63} - Valid bug, valid fix: known issue in SQLite \url{https://stackoverflow.com/questions/28385069/sqliteopenhelper-setwriteaheadloggingenabled-causes-an-error-log-line}. 
 
    \item * - \url{https://github.com/mpcjanssen/ubiquitous-capture/issues/4} - Valid bug, valid fix. Not idenfitiable because this is fundamentally a UX bug.
 
    \item \url{https://github.com/assertj/assertj-vavr/issues/141} - Valid bug, valid fix. Identifiable based on other patterns in the same repository. 
 
    \item \url{https://github.com/smartdevicelink/sdl_java_suite/issues/53} - Valid issue, valid fix. Identifiable through other similar patterns in similar code in the repository.
 
    \item \url{https://github.com/darcy-framework/darcy-webdriver/issues/30} - Valid bug, valid fix. Identifiable through other similar patterns in similar code in the repository.
 
    \item * - \url{https://github.com/AlexFalappa/nb-springboot/issues/167} - Valid bug, valid fix. Not identifiable.
 
    \item \url{https://github.com/GabrielOlvH/Carrier/issues/2} - Valid issue, fix not permanent, but does indeed correctly point to the location of the problem. The problem, broadly speaking, is caused by the fact that the Wolf entity is different from all of the other entities, and therefore calling \texttt{updateHolding} method on it will cause issues.
 
    \item \url{https://github.com/AgriCraft/AgriCraft/issues/82} - Valid issue, valid fix. It’s logic that is identifiable by humans by looking at the variable names and intended usage.
 
    \item \url{https://github.com/rasmus-saks/aken-ajalukku/issues/65} - Valid issue, valid fix. Identifiable based on the context of the application, the fact that this is actually a “walking tour”, therefore the mode on google maps should be for “walking” rather than driving.
 
    \item \url{https://github.com/CJMinecraft01/DoubleSlabs/issues/81} - Valid issue, valid fix. Fix is identifiable in principle, but a lot of context about how minecraft slabs interact is needed.
 
    \item \url{https://github.com/hzi-braunschweig/SORMAS-Project/issues/6832} - Valid issue, valid fix. Bug is identifiable, the start date is replaced with Enddate in some cases on records.
 
    \item \url{https://github.com/MachinePublishers/jBrowserDriver/issues/67} - Valid issue, valid fix. The bug is identifiable in the long context and with knowledge of the general cookie-creation pattern.  The problem is that the domain for the cookie is not set, so it’s not being used by the web-driver on repeat visits to a website.
 
    \item \url{https://github.com/release-engineering/pom-manipulation-ext/issues/240} - Valid issue, valid fix. This one just fixes an NPE, but it does contain a lot of style/whitespace changes.
 
    \item \url{https://github.com/Angry-Pixel/The-Betweenlands/issues/948} - Valid bug, valid fix. This one is in principle identifiable with knowledge of the pattern in minecraft servers to have different types of blocks that constitute a single “entity” (a door in this case). 
 
    \item \url{https://github.com/ehcache/ehcache3/issues/2638} - Valid bug, valid fix. Bug is straightforward and  identifiable.
 	
    \item \url{https://github.com/thingsboard/thingsboard/issues/3992} - Valid bug, valid fix. The essence is that the method \texttt{getDeviceTypes} should call the “/devices/types” api endpoint rather than “/devices”. Should be identifiable based on semantic context.
 
    \item \url{https://github.com/vert-x3/vertx-config/issues/20} - Valid bug, valid fix. The bug is easily identifiable because (a) Vertx is often used in code, and (b) the “host” variable is left unused despite being declared, and there is only one logical place where it can be potentially used.
 
    \item \url{https://github.com/metarhia/jstp-java/issues/24} - Valid bug, valid fix. The bug is identifiable given context about the \texttt{ExecutionHandler} class.
 
    \item * - \url{https://github.com/Cassiobsk8/Industrial\_Renewal/issues/126} - Valid bug, valid fix. This bug is diffichttps://www.overleaf.com/project/63e2c9d0daa4bbc401a23fbault to identify because (a) the fix is to override a particular method of the class and (b) it’s not obvious that there can be an obstruction with bunk beds..
 
    \item \url{https://github.com/Tamaized/AoV/issues/13} - Valid bug, valid fix. Contains some whitespace additions in addition to bug fix. Bug related to a certain config option not being used, it’s identifiable through semantic understanding of the code.
 
    \item * - \url{https://github.com/PyvesB/advanced-achievements/issues/172} - Valid bug, valid fix. The bug is hard to identify, because it requires context about brewing stand operations. 
 
    \item \url{https://github.com/eclipse/vorto/issues/442} - Valid bug, valid fix. The issue is identifiable by the fact that resource id is hardcoded to “0” rather than the “resourceId” variable provided. Also semantically identifiable. Contains whitespace changes as well.
 
    \item \url{https://github.com/hsyyid/AdminShop/issues/5} - Valid bug, valid fix. Identifiable bug.
 
    \item * - \url{https://github.com/labhackercd/edm/issues/5} - Valid bug, unsure if valid fix. The bug has little information, and the crash that the bug reports does not seem to be identifiable from the code alone (maybe it’s device-dependent, but the pattern used that is considered buggy is widely recommended, see \url{https://stackoverflow.com/questions/2422562/how-to-change-theme-for-alertdialog)}.
 
    \item \url{https://github.com/Haptic-Apps/Slide/issues/655} - Valid bug, valid fix. A run-time exception from a method should be catched.
 
    \item * - \url{https://github.com/PortuguesDoSeculoXXI/PortuguesDoSeculoXXI/issues/48} - Unsure, hard to ascertain, there is a lot of text in portuguese
 
    \item \url{https://github.com/OpenJEVis/JEVis/issues/840} - Valid bug, valid fix. Although the text is in German, the bug is about processing a list of items into a menu rather than just one. This is not a new “feature”, since the items to be processed were always packaged in a variable-length “list”.
 
    \item \url{https://github.com/commons-app/apps-android-commons/issues/587} - Valid bug, valid fix. Links need to be sanitized.
 
    \item \url{https://github.com/twizmwazin/CardinalPGM/issues/86} - Valid bug, valid fix. A map cycle schedule was only set when time was under 5 seconds, so the scheduling needed to be moved out of that particular if statement.
 
    \item * - \url{https://github.com/assemblits/eru/issues/100} - Not a valid bug. The change is just a change to the title of an alert to set it to a class name rather than a generic “connection failure” message.
 
    \item \url{https://github.com/MachinePublishers/jBrowserDriver/issues/21} - Valid bug, valid fix. The issue is that cookies come in a lot of formats, and not all of them were supported by jBrowser. 
 
    \item \url{https://github.com/lucas-tulio/server-simulator/issues/6} - The bug is valid, fix invalid.
 
    \item \url{https://github.com/spring-cloud/spring-cloud-netflix/issues/1724} - Valid bug, valid fix. The property for \texttt{preferIPAddress} was not taken into account. The config is needed to be able to identify this issue.
 
    \item \url{https://github.com/Col-E/Recaf/issues/344} - Valid bug, valid fix. The previous version replaced all \texttt{\$} in a class name with \texttt{.}, but only the last one needs to be replaced by convention.
 
    \item \url{https://github.com/plan-player-analytics/Plan/issues/1313} - Valid bug, valid fix. The front-end called the wrong endpoint, and the backend was adjusted so that the front-end was calling the correct endpoint to get a list of players for a server.
 
    \item * - \url{https://github.com/cabaletta/baritone/issues/330} - Not really a bug, more like an extra feature. Also has the “enhancement” tag. 
 
    \item \url{https://github.com/sosy-lab/java-common-lib/issues/19} - Valid bug, valid fix. The problem was that one of the iterators in a method that returned sorted list of two collections wasn’t fully exhausted.
 
    \item \url{https://github.com/Cactiw/Timetable/issues/4} - Valid bug, valid fix. The issue was that an update task wasn’t put into an async call, and therefore caused issues downstream. Putting it in async fixes the issue.
 
    \item \url{https://github.com/jooby-project/jooby/issues/1489} - Valid bug, valid fix. Pretty easily identifiable that the factory is closed rather than the session that was just checked in the surrounding if statement.
 
    \item \url{https://github.com/Gaming32/ArrayV-v4.0/issues/43} - Valid bug, valid fix. Method that obviously should have been \texttt{synchronized} based on surrounding code wasn't \texttt{synchronized}.
 
    \item \url{https://github.com/Zedd7/ZHorse/issues/25} - Valid bug, valid fix. The duplicate horse is not assigned a name, so when “deleting” it the message should display the original horses’s name.
 
    \item \url{https://github.com/neo4j-contrib/neo4j-apoc-procedures/issues/303} - Valid bug, valid fix. The issue is that two nodes cannot have the same “main key”, and when merging two nodes the previous node was not deleted therefore causing an exception. The solution is that the properties of the source node should be stored, the source node deleted, and then the properties of the source node have to loaded into the target node from the stored variable.
 
    \item \url{https://github.com/mikepenz/CrossfadeDrawerLayout/issues/15} - Valid bug, valid fix. The bug is that when opening/closing the drawer, the state is not necessarily updated. The fix is to override the appropriate methods to update state. Identifiable by the common pattern of state updates when calling certain parent class methods.
 
    \item \url{https://github.com/wultra/powerauth-webflow/issues/345} - Valid bug, valid fix. The issue is that the message of the logger wasn’t aligned with the exception being caught. There are also some whitespace changes added in the commit.
 
    \item \url{https://github.com/TeamLapen/Vampirism/issues/333} - Valid bug, valid fix. Off-by-one error, identifiable by knowing about minecraft item stacks as well as generally looking around the code.
 	
    \item \url{https://github.com/home-climate-control/dz/issues/144} - Valid bug, valid fix. Authentication parameters passed to the bean were not actually set when creating the bean. Bug is identifiable.
 
    \item \url{https://github.com/kontalk/androidclient/issues/1264} - Valid bug, valid fix. The problem is that the push notification service would not be started as a foreground service and the system would kill it after 15 seconds. The fix is to start it in the foreground before. Identifiable because this is a common pattern in messaging applications.
 
    \item \url{https://github.com/BetonQuest/BetonQuest/issues/734} - Valid bug, valid fix. The problem is that the method used previously to detect entity deaths in Minecraft was sub-optimal, and it was replaced by a better method that was seen in a different minecraft plugin. Identifiable with knowledge of the spigot library usage patterns.
 
    \item \url{https://github.com/jmockit/jmockit1/issues/98} - Valid bug, valid fix. The problem is that sometimes types that should be \texttt{null} are not mocked as \texttt{null} objects. The fix addresses these cases with cascading types. Bug is identifiable with knowledge mocking patterns.
 
    \item * - \url{https://github.com/Freeyourgadget/Gadgetbridge/issues/529} - Invalid issue that has been deleted. 
 
    \item \url{https://github.com/scenerygraphics/sciview/issues/181} - Valid bug, valid fix. One needs to call \texttt{setSize} on the panel before displaying it, and the fix addresses that.
 
    \item \url{https://github.com/ramack/ActivityDiary/issues/153} - Valid bug, valid fix. Identifiable through app context.
 
    \item \url{https://github.com/decarbonization/android-fonz/issues/26} - Valid bug, valid fix. Type mismatch in settings crashed the app. Identifiable from android property conventions.
 
    \item \url{https://github.com/ICIJ/datashare/issues/41} - Valid bug, valid fix. Fix the ISO code representations of languages by a) adding more enums and b) using both iso1 and iso2 codes to identify a language. Should be easily identifiable since both iso1 and iso2 parameters are passed in.
 
    \item \url{https://github.com/twizmwazin/CardinalPGM/issues/645} - Valid bug, valid fix. The fix is to use a more high-level API provided by the library in question, and simplify the existing code dramatically and add support for detecting TNT damage and account for points. Should be identifiable given the whole library context.
 
    \item * - \url{https://github.com/manoelcampos/cloudsimplus/issues/368} - Valid bug, valid fix. This is an issue that would be hard to identify since it’s configuration/resource usage based, and depends on system parameters a lot.
 
    \item * - \url{https://github.com/almosr/android-svg-code-render/issues/67} - Valid bug, valid fix. Not identifiable due to the bug being in templating.
 
    \item \url{https://github.com/dcm4che/dcm4chee-arc-light/issues/523} - Valid bug, valid fix. When a study is deleted, the number of studies that patients of this study participated in has to be decreased. Identifiable with understanding of relationship between patients and studies.
 
    \item * - \url{https://github.com/BCA-Team/Buildcraft-Additions/issues/356} - Valid bug, valid fix. Minecraft lasers fired underwater get stuck. The fix implements the logic of dissipating the laser once it hits lava or water. Not identifiable without knowing in-app logic about lasers and expected behavior upon hitting water or lava.
 
    \item \url{https://github.com/TotalHamman/BetterBlockExchanger/issues/7} - Valid bug, valid fix. The app was reading state from the “previous” state during a swap rather than the “current” state, causing an NPE.
 
    \item \url{https://github.com/BasicAirData/GPSLogger/issues/132} - Valid bug, valid fix. The contents of a variable on which a switch was conditioned could be potentially null, and that caused an NPE.
 
    \item \url{https://github.com/OfficeDev/ews-java-api/issues/8} - Valid bug, valid fix. The bug is identifiable by the fact that there is an used variable, and the only place where it can be reasonably used is in a method override of a method inherited from the parent class.
 
    \item \url{https://github.com/dcm4che/dcm4chee-arc-light/issues/1180} - Valid bug, valid fix. The problem is that the program assigns the “type” of a SOP instance not according to the DICOM specification. Knowledge of the DICOM specification is necessary to identify the bug.
 
    \item \url{https://github.com/LMBishop/Quests/issues/281} - Valid bug, valid fix. The problem is that player quests are not restored to the player object once the player joined the server. Identifiable with common quest/server patterns.
 
    \item * - \url{https://github.com/danielricci/solitaire/issues/90} - Valid bug, invalid fixes.
 
    \item \url{https://github.com/hv0905/SchoolStoryCollection/issues/2} - Valid bug, valid fix, identifiable by human.
 
    \item \url{https://github.com/davidcorbin/mygcc-api/issues/21} - Valid bug, valid fix. The issue is that when processing an image the code searches for the last occurrence of “.jpg”, but this cannot be found and throws an error if the image URL is uppercase.
 
    \item \url{https://github.com/Tamaized/AoV/issues/104} - Valid bug, valid fix. The issue is that a player can just hop in and out of bed to recharge certain abilities, but in reality they need to fully sleep in the bed for that. Identifiable bug because this is a common pattern (you need to actually \textit{sleep}) in many minecraft games.
 
    \item \url{https://github.com/Electroblob77/Wizardry/issues/513} - Valid bug, valid fix. The fix consists of using the stream and filter apis to avoid removing items from a list that can be concurrently modified. Identifiable with knowledge of concurrency mechanisms.
 
    \item \url{https://github.com/tsandmann/ct-sim/issues/62} - Valid bug (because the code does not correspond to the documentation comments), valid fix. Interestingly the documentation is in German.
 
    \item \url{https://github.com/voxelwind/voxelwind/issues/33} - Valid bug, valid fix. The problem is that the project implements a server for Minecraft: Pocket Edition, and it doesn’t fully comply with the API contract, in particular with respect to the yaw parameters that clients pass in. Identifiable with knowledge of the API.
 
    \item \url{https://github.com/phrack/ShootOFF/issues/651} - Valid bug, valid fix. Wrong class was used to test if a shot color matched certain constants. Identifiable.
 
    \item * - \url{https://github.com/rmichela/GiantTrees/issues/31} - Not really a bug, this just implements mechanisms that silence warnings if certain resource files do not exist.
 
    \item \url{https://github.com/sriharshachilakapati/SilenceEngine/issues/38} - Valid bug, valid fix. If the vertices for a certain polygon are cleared, some parameters are set to maximum and minimum infinity, and some downstream methods throw errors. Fix is modification of these methods to account for cases when vertices are 0.
 
    \item \url{https://github.com/lsfusion/platform/issues/164} - Valid bug, valid fix. The method would not account for ftp files that did not exist, and the modification allows the method to handle non-existent ftp files as well.
 
    \item \url{https://github.com/spring-cloud/spring-cloud-sleuth/issues/1816} - Valid bug, valid fix. The sleuth library was interfering with openfeign’s circuitbreaker capabilities, the fix was to conditionally create the feign bean only if circuitbreaker was disabled. Identifiable by humans.
 
    \item \url{https://github.com/vinaygaba/CreditCardView/issues/13} - Valid bug, valid fix. The issue is that the setter methods did not modify the actual state of the class. Easily identifiable by humans.
 
    \item \url{https://github.com/AludraTest/aludratest/issues/36} - Valid bug, valid fix. Identifiable with context of the Selenium library.
 
    \item \url{https://github.com/VazkiiMods/Quark/issues/3374} - Valid bug, valid fix. Simple fix to fix the formatting of chat events in certain cases of item links. 
 
    \item \url{https://github.com/Tamaized/AoV/issues/93} - Valid bug, valid fix. The problem is that casting the furious howl spell should only apply to a selected target. Identifiable through other examples of spells cast on a specific target in the library.
 
    \item \url{https://github.com/spring-projects/spring-boot-data-geode/issues/55} - Valid bug, valid fix. Identifiable with knowledge of Spring beans.
 
    \item \url{https://github.com/twizmwazin/CardinalPGM/issues/657} - Valid bug, valid fix. Identifiable through Bukkit/game conventions: the game sets player metadata, but doesn’t remove it on match end. 
 
    \item \url{https://github.com/rundeck/rundeck-cli/issues/43} - Valid bug, valid fix. Annotation text doesn’t align with documentation about command line option usage in the rest of the code.

\end{enumerate}

\section{Synthetic Dataset Samples}
\label{appendix:synthetic_samples}

\subsection{Example of InCoder perturbation}
We present an example of a sample perturbation generated by the InCoder model in Figure \ref{fig:perturbed}. The original observation is on the left-hand side, and the perturbed observation is on the right-hand side.

\begin{figure*}[h!]
\centering
\setlength{\tabcolsep}{5pt}
\begin{tabular}{@{}p{0.49\linewidth}p{0.49\linewidth}@{}}
\begin{lstlisting}[basicstyle=\tiny, language=Java, linebackgroundcolor={%
    % \btLstHL{10-17}
    \ifnum\value{lstnumber}=10\color{ll1}\fi
    \ifnum\value{lstnumber}=11\color{ll1}\fi
    \ifnum\value{lstnumber}=12\color{ll1}\fi
    \ifnum\value{lstnumber}=13\color{ll1}\fi
    \ifnum\value{lstnumber}=14\color{ll1}\fi
    \ifnum\value{lstnumber}=15\color{ll1}\fi
    \ifnum\value{lstnumber}=16\color{ll1}\fi
    \ifnum\value{lstnumber}=17\color{ll1}\fi
}]
package com.github.yt.mybatis.utils;

import com.github.yt.base.exception.BaseErrorException;
import org.apache.commons.lang3.StringUtils;
import java.lang.reflect.Field;

public class BeanUtils {

    public static ChainMap<String, Object>
      getValueMap(Object... objs) {
        try {
            ChainMap<String, Object> map =
      new ChainMap<>();
            for (Object obj : objs) {
                if (null == obj) {
                    continue;
                }
                for (Class<?> c = obj.getClass(); Object.class!= c; c = c.getSuperclass()) {
                    for (Field field : c.getDeclaredFields()) {
                        field.setAccessible(true);
                        Object value = field.get(obj);
                        if (null == value) {
                            continue;
                        }
                        if (field.getType().isAssignableFrom(String.class) && StringUtils.isEmpty((String) value)) {
                            continue;
                        }
                        map.put(field.getName(), value);
                    }
                }
            }
            return map;
        } catch (Exception e) {
            throw new BaseErrorException("Object to Map convert Error", e);
        }
    }
}
  \end{lstlisting}&
  \begin{lstlisting}[basicstyle=\tiny,language=Java,linebackgroundcolor={%
    \ifnum\value{lstnumber}=10\color{ll2}\fi
    \ifnum\value{lstnumber}=11\color{ll2}\fi
    \ifnum\value{lstnumber}=12\color{ll2}\fi
    \ifnum\value{lstnumber}=13\color{ll2}\fi
    \ifnum\value{lstnumber}=14\color{ll2}\fi
    \ifnum\value{lstnumber}=15\color{ll2}\fi
    \ifnum\value{lstnumber}=16\color{ll2}\fi
}]
package com.github.yt.mybatis.utils;

import com.github.yt.base.exception.BaseErrorException;
import org.apache.commons.lang3.StringUtils;
import java.lang.reflect.Field;

public class BeanUtils {

    public static ChainMap<String, Object>
      getFieldMap(Object... objs) {

        ChainMap<String, Object> map =
      new ChainMap<String, Object>();
        try {
            if (objs != null && objs.length == 1) {
                Object obj = objs[0];
                for(Class<?> c = obj.getClass(); Object.class!= c; c = c.getSuperclass()) {
                    for (Field field : c.getDeclaredFields()) {
                        field.setAccessible(true);
                        Object value = field.get(obj);
                        if (null == value) {
                            continue;
                        }
                        if (field.getType().isAssignableFrom(String.class) && StringUtils.isEmpty((String) value)) {
                            continue;
                        }
                        map.put(field.getName(), value);
                    }
                }
            }
            return map;
        } catch (Exception e) {
            throw new BaseErrorException("Object to Map convert Error", e);
        }
    }
}
  \end{lstlisting}\\
  \vspace{-10pt}
  \caption*{Original observation.} & \vspace{-10pt} \caption*{Perturbed observation.}
 \end{tabular}
 \vspace{-10pt}
 \caption{Sampled perturbation introduces a non-trivial rewrite, which may be considered as a ``bug''. \textit{Left}: Original Java code iterates over a given list of objects (green highlight). \textit{Right}: Perturbed Java code only considers the first object in the list, if the list contains precisely one element (red highlight).}
 \label{fig:perturbed}
\end{figure*}

Remarkably, both sequences appear to be syntactically correct code. The auto-repressive sampler took future tokens into account. For example, the type resolution of the object \inlinecode{map} may be resolved by the return signature of the function \inlinecode{public static ChainMap<...>} which was not masked out and the invocation of \inlinecode{map.put(...)}. While the original code iterates over the list of objects \inlinecode{obj}, the perturbed code only considers the first element of the list, if the list contains a single element. Whether the rewrites constitute a "bug" depends on the definition of the term, as earlier discussed. However, given the context one can argue that the rewritten implementation seems less probable to follow the underlying intent.

\subsection{Synthetic Samples with Explanations}
In this section we provide a few samples of bugs generated in the second revision of the synthetic dataset as well as brief explanations of the perturbations in Figures \ref{fig:sample_1}, \ref{fig:sample_2}, and \ref{fig:sample_3}. The red lines mark lines affected by the perturbations.

In Sample 1 (\ref{fig:sample_1}), the bug is obviously introduced because the password string is being printed out into STDOUT. This is valid Java code, bug a significant deviation from secure coding practices. The code in Sample 2 ( \ref{fig:sample_2}) endless recursion loop call. The code in Sample 3 (\ref{fig:sample_3}) is from a Spring app that controls a Planet API. Every time a call to \texttt{addPlanet} API is made, the API returns \texttt{HttpStatus.BAD\_REQUEST} every time regardless of whether the operation succeeded or not. 

\begin{figure}[h!]
    \begin{lstlisting}[basicstyle=\tiny,language=Java,linebackgroundcolor={%
    \ifnum\value{lstnumber}=16\color{ll2}\fi
    \ifnum\value{lstnumber}=17\color{ll2}\fi
    \ifnum\value{lstnumber}=18\color{ll2}\fi
    \ifnum\value{lstnumber}=19\color{ll2}\fi
    \ifnum\value{lstnumber}=20\color{ll2}\fi
    \ifnum\value{lstnumber}=22\color{ll2}\fi
}]
      @Override
      protected AuthenticationInfo doGetAuthenticationInfo(AuthenticationToken token) throws AuthenticationException {
        System.out.println("==登录认证==");
        //第一步：从token中取出用户名
        String realname = (String)token.getPrincipal();

        // 第二步：根据用户名 从数据库 获取用户
        User user = null;
        try {
          user = userService.findObjectByName(realname);
        } catch (Exception e1) {
             // TODO Auto-generated catch block
             e1.printStackTrace();
         }
         if(user == null) {
             return  new SimpleAuthenticationInfo("用户不存在!", false, getName());
         }
         //密码
         String password = getPasswordEncoder().encodeToString(user.getPassword());
         System.out.println("认证内"+password);
 
         //第三步：根据用户名�返回认证信息AuthenticationInfo
 
         //activeUser就是用户身份信息
         ActiveUser activeUser = new ActiveUser();
         activeUser.setRealname(user.getRealname());
         activeUser.setPhone(user.getPhone());
         activeUser.setValid(user.getValid());
 
 
         //System.out.println("认证内"+activeUser);
         //自动完成密码比对   - 密码的比对:
         //通过 AuthenticatingRealm 的 credentialsMatcher 属性来进行的密码的比对!
         SimpleAuthenticationInfo info =
                 new SimpleAuthenticationInfo(activeUser,password,credentialsSalt,getName());
         SecurityUtils.getSubject().getSession().setAttribute("currentUser",user);
         return info;
     }
 }
  \end{lstlisting}
  \caption{Revised Synthetic Dataset Sample 1}
  \label{fig:sample_1}
\end{figure}

\begin{figure}[h!]
    \begin{lstlisting}[basicstyle=\tiny,language=Java,linebackgroundcolor={%
    \ifnum\value{lstnumber}=18\color{ll2}\fi
    \ifnum\value{lstnumber}=19\color{ll2}\fi
    \ifnum\value{lstnumber}=20\color{ll2}\fi
    \ifnum\value{lstnumber}=22\color{ll2}\fi
    \ifnum\value{lstnumber}=23\color{ll2}\fi
    \ifnum\value{lstnumber}=24\color{ll2}\fi
    \ifnum\value{lstnumber}=25\color{ll2}\fi
    \ifnum\value{lstnumber}=26\color{ll2}\fi
    \ifnum\value{lstnumber}=27\color{ll2}\fi
}]
   public class LazyFragment extends Fragment {
 
      protected LayoutInflater inflater;
      private View contentView;
      private Context context;
      private ViewGroup container;
 
      @Override
      public void onCreate(Bundle savedInstanceState) {
         super.onCreate(savedInstanceState);
         context = getActivity().getApplicationContext();
      }
 
      @Override
      public  View onCreateView(LayoutInflater inflater, ViewGroup container, Bundle savedInstanceState) {
         this.inflater = inflater;
         this.container = container;
         onCreateiew(inflater, container, savedInstanceState);
         return contentView;
      }
 
      @Override
      public void onDestroyView() {
         super.onDestroyView();
         contentView = null;
         container = null;
         inflater = null;
      }
  \end{lstlisting}
  \caption{Revised Synthetic Dataset Sample 2}
  \label{fig:sample_2}
\end{figure}

\begin{figure}[h!]
    \begin{lstlisting}[basicstyle=\tiny,language=Java,linebackgroundcolor={%
    \ifnum\value{lstnumber}=36\color{ll2}\fi
    \ifnum\value{lstnumber}=37\color{ll2}\fi
    \ifnum\value{lstnumber}=38\color{ll2}\fi
    \ifnum\value{lstnumber}=39\color{ll2}\fi
    \ifnum\value{lstnumber}=40\color{ll2}\fi
    \ifnum\value{lstnumber}=41\color{ll2}\fi
    \ifnum\value{lstnumber}=42\color{ll2}\fi
    \ifnum\value{lstnumber}=43\color{ll2}\fi
    \ifnum\value{lstnumber}=44\color{ll2}\fi
    \ifnum\value{lstnumber}=45\color{ll2}\fi
    \ifnum\value{lstnumber}=46\color{ll2}\fi
    \ifnum\value{lstnumber}=47\color{ll2}\fi
}]
 @Controller
 @RequestMapping(value = "/api")
 @CrossOrigin(origins = "*")
 //@CrossOrigin(origins = "http://localhost:4200")
 public class PlanetController {
 	
 //	private PlanetService planetService; I DONT HAVE A SERVICE AT THE MOMENT
 	
 	private PlanetRepo planetRepo;
 
 	public PlanetController(PlanetRepo planetRepo) {
 		super();
 		this.planetRepo = planetRepo;
 	}
 	
 	
 //	@RequestMapping(value = "/getPlanets", method = RequestMethod.GET)
 	@GetMapping(value = "/getPlanets") //These are identical in configuration. http://localhost:8080/SpringMVC/mvc/api/getPlanets
 	public @ResponseBody List<Planet> getAllPlanets(){
 		return planetRepo.selectAll();
 	}
 	
 	@GetMapping(value = "/getPlanetsWithParam", produces="application/json", params = {"id"}) //http://localhost:8080/SpringMVC/mvc/api/getPlanetsWithParam?id=2
 	public ResponseEntity<Planet> getPlanetById(int id) {
 		return new ResponseEntity<Planet>(planetRepo.selectPlanetById(id), HttpStatus.INSUFFICIENT_STORAGE);
 	}
  	
 	
 	@ResponseStatus(value = HttpStatus.ALREADY_REPORTED)
 	@GetMapping(value = "/getPlanetWithUrl/{num}")
 	public @ResponseBody Planet getPlanetByIdTwo(@PathVariable("num") int id ) {
 		return planetRepo.selectPlanetById(id);
 	}
 	
 	
 	@ResponseStatus(value = HttpStatus.BAD_REQUEST)
 	@PostMapping(value = "/addPlanet")
 	public @ResponseBody String addPlanet(@RequestBody Planet incomingPlanet) {
 		
 		/*
 		 * This method is executed when the user requests to add a new planet to 
 		 * the database. 
 		 * 
 		 * The default behavior for MVC is to ignore the incoming JSON 
 		 * and treat it like if it were a GET request. 
 		 * 
 		 * In our example, we would expect this request to  If the incoming JSON does NOT HAVE all the fields, it will provide just default values. 
 		 */
 		planetRepo.insert(incomingPlanet);
 		
 		return "Success";
 	}

 	@GetMapping(value = "/allTheHeaders")
 	public ResponseEntity<String> allHeaders(@RequestHeader Map<String,String> allHeaders){
 		
 		//THIS IS NOTHING TO DO WITH MVC
 		//This is from Collections (Week 1)
 		for(Entry<String, String> entry: allHeaders.entrySet()) {
 			System.out.println(entry.getKey() + "\t" + entry.getValue());
 		}
 		
 		HttpHeaders responseHeader = new HttpHeaders();
 		
 		responseHeader.set("Name", "Bobby");
 		responseHeader.set("superSecrets", "********");
 		
 		return new ResponseEntity<String>("Success", responseHeader, HttpStatus.FORBIDDEN);
 	}
 
 }
  \end{lstlisting}
  \caption{Revised Synthetic Dataset Sample 3}
  \label{fig:sample_3}
\end{figure}

\section{Realistic Benchmark Samples}
\label{appendix:realistic_samples}

In order to show the necessity for long-context language models for bug localization, we demonstrate an example of a bug that is highly dependent on external context outside of the scope of the file where the bug is located. The issue \footnote{https://github.com/mewna/catnip/issues/105} in question is related to a bug in a software project Catnip that provides a Discord\footnote{https://discord.com} API wrapper in Java. The bug is that the Java plugin uses \texttt{java.awt.Color}\footnote{https://docs.oracle.com/javase/7/docs/api/java/awt/Color.html} to store Color parameter for embeds. This class stores not only the RGB bits of the color, but also 8 extra alpha bits, constituting a 64 bit representation of each color. However, the Discord API only accepts 48 bit representations of colors. The fix, therefore, consists of masking out the first 8 bits of the color representation of \texttt{java.awt.Color} and passing that into the API. The fixing hunk is shown in Figure \ref{fig:diff}.

For a human to be able to understand and identify this issue, they need to know about the Discord API, the common patterns of calling the Discord API, the peculiarities of color representations in the \texttt{java.awt.Color} class, and how to perform bit operations. We present an piece of code \footnote{https://discordjs.guide/popular-topics/embeds.html\#embed-preview} available online that demonstrates calling the discord API. This code passes in a color for the embed frame as a 48-bit digit (Figure \ref{fig:issue_example}). Examples like this in training or in the context are necessary for the model to have a chance at locating this bug. Without the context, even a human observer cannot reliably mark this as buggy code.



\begin{figure}[h!]
    \begin{lstlisting}[basicstyle=\tiny,language=Java,linebackgroundcolor={%
    \ifnum\value{lstnumber}=4\color{ll2}\fi
    \ifnum\value{lstnumber}=5\color{ll1}\fi
    \ifnum\value{lstnumber}=6\color{ll1}\fi
}]
    @CheckReturnValue
    public EmbedBuilder color(@Nullable final Color color) {
        if (color != null) {
            this.color = color.getRGB();
            // Mask off the alpha bits
            this.color = color.getRGB() & 0x00FFFFFF;
        }
        return this;
    }
  \end{lstlisting}
  \caption{Hunk from sample issue from Catnip. This bug demonstrates the need for more context than file-level information.}
  \label{fig:diff}
\end{figure}

\begin{figure}[h!]
    \begin{lstlisting}[basicstyle=\tiny,linebackgroundcolor={%
    \ifnum\value{lstnumber}=2\color{lightyellow}\fi
}]
    const exampleEmbed = new EmbedBuilder()
	    .setColor(0x0099FF)
	    .setTitle('Some title')
	    .setURL('https://discord.js.org/')
	    .setAuthor({ name: 'Some name',
	        iconURL: 'https://i.imgur.com/AfFp7pu.png', 
	        url: 'https://discord.js.org' })
	    .setDescription('Some description here')
	    .setThumbnail('https://i.imgur.com/AfFp7pu.png')
	    .addFields(
		    { name: 'Regular field title', value: 'Some value here' },
		    { name: '\u200B', value: '\u200B' },
		    { name: 'Inline field title', value: 'Some value here', inline: true },
		    { name: 'Inline field title', value: 'Some value here', inline: true },
	    )
	    .addFields({ name: 'Inline field title', value: 'Some value here', inline: true })
	    .setImage('https://i.imgur.com/AfFp7pu.png')
	    .setTimestamp()
	    .setFooter({ text: 'Some footer text here', iconURL: 'https://i.imgur.com/AfFp7pu.png' });
  \end{lstlisting}
  \caption{Example of calling the Discord API with a RGB (24-bit) color representation while RGBA (32-bit) is expected}
  \label{fig:issue_example}
\end{figure}

\end{document}